\pdfoutput=1

\documentclass[11pt]{article}

\usepackage[preprint]{latex/acl}
\usepackage{times}
\usepackage{latexsym}

\usepackage[T1]{fontenc}

\usepackage[utf8]{inputenc}

\usepackage{microtype}

\usepackage{inconsolata}

\usepackage{graphicx}
\usepackage{amsmath,amssymb,amsfonts}
\usepackage{booktabs}
\usepackage{graphicx} 
\usepackage{multirow}
\usepackage{cleveref}
\usepackage{float}
\usepackage{placeins}
\title{RetrieveAll: A Multilingual Named Entity Recognition Framework with Large Language Models}

\author{
  \textbf{Jin Zhang\textsuperscript{1}\thanks{Equal contribution.}},
  \textbf{Fan Gao\textsuperscript{2}\footnotemark[1]},
  \textbf{Linyu Li\textsuperscript{3}}\\
  \textbf{Yongbin Yu\textsuperscript{2}\thanks{Corresponding author.}},
  \textbf{Xiangxiang Wang\textsuperscript{2}\footnotemark[2]},
  \textbf{Nyima Tashi\textsuperscript{1}\footnotemark[2]},
  \textbf{Gadeng Luosang\textsuperscript{1}\footnotemark[2]} \\
\\
  \textsuperscript{1} School of Information Science and Technology, Tibet University \\
  \textsuperscript{2}School of Information and Software Engineering,\\ University of Electronic Science and Technology of China \\
  \textsuperscript{3}School of Computer Science, Peking University \\
  \texttt{jinzhang@stu.utibet.edu.cn}, 
  \texttt{202221090302@std.uestc.edu.cn} 
}

\begin{document}
\maketitle
\begin{abstract}
The rise of large language models has led to significant performance breakthroughs in named entity recognition (NER) for high-resource languages, yet there remains substantial room for improvement in low- and medium-resource languages. Existing multilingual NER methods face severe language interference during the multi-language adaptation process, manifested in feature conflicts between different languages and the competitive suppression of low-resource language features by high-resource languages. Although training a dedicated model for each language can mitigate such interference, it lacks scalability and incurs excessive computational costs in real-world applications.  To address this issue, we propose RetrieveAll, a universal multilingual NER framework based on dynamic LoRA. The framework decouples task-specific features across languages and demonstrates efficient dynamic adaptability. Furthermore, we introduce a cross-granularity knowledge augmented method that fully exploits the intrinsic potential of the data without relying on external resources. By leveraging a hierarchical prompting mechanism to guide knowledge injection, this approach advances the paradigm from “prompt-guided inference” to “prompt-driven learning.”
Experimental results show that RetrieveAll outperforms existing baselines; on the PAN-X dataset, it achieves an average F1 improvement of 12.1\%.
\end{abstract}

\section{Introduction}
\begin{figure}[t]
  \centering
  \includegraphics[width=\columnwidth]{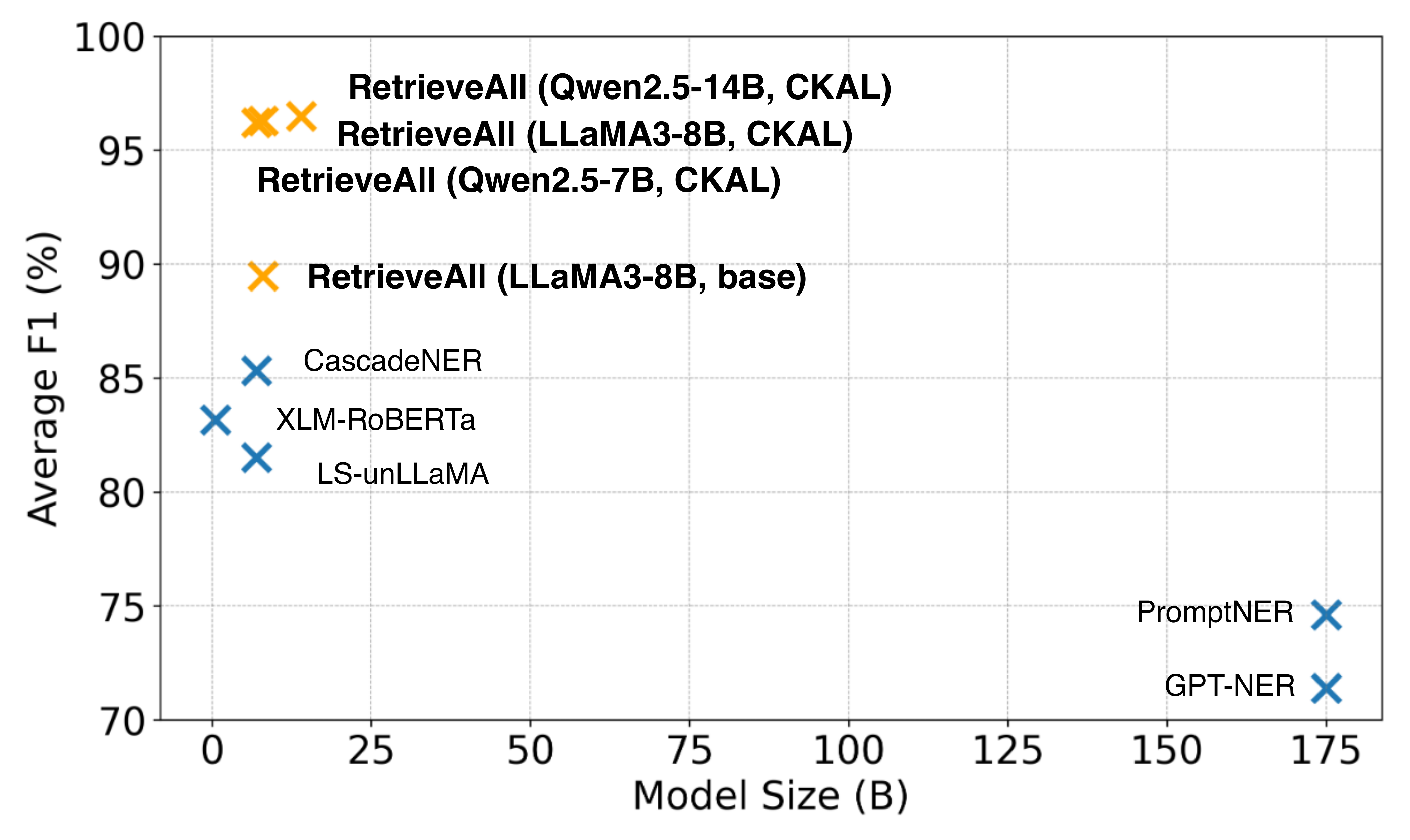}  
  \caption{The figure illustrates the relationship between model size and average F1 scores across eight representative languages (English, Spanish, French, Russian, German, Chinese, Japanese, and Korean) on the PAN-X dataset for RetrieveAll. RetrieveAll consistently delivers substantial performance gains across different base models, combining efficiency with outstanding results.}  
  \label{main}  
\end{figure}

Named Entity Recognition (NER) has gradually become an important research direction in the field of Natural Language Processing (NLP), playing a crucial role in downstream applications such as knowledge graph construction and reasoning \citep{zhu2024llms}, event extraction \citep{chen2024large}, and question answering systems \citep{alan2024rag}. The main goal of NER is to accurately identify the entities mentioned in a sentence and classify them into the corresponding entity types. Existing NER methods for high-resource languages have achieved significant performance breakthroughs through large language models (LLMs). However, there is still considerable room for improvement in NER for medium- and low-resource languages. Multilingual NER faces a series of complex challenges.

Existing multilingual NER approaches can be grouped into two directions: one involves training a special model for each individual language, while the other employs a unified model capable of recognizing multiple languages simultaneously. 
The approach of training a specific model for each language does not scale effectively, as a new model needs to be trained and stored every time a new language is introduced, resulting in significant time and storage consumption.

When employing a unified model to handle multiple languages, where the model concurrently learns data from various languages, a critical challenge that arises is "language interference". Language interference refers to the potential negative impact on the model's performance on certain languages due to the model learning the characteristics of other languages. Language interference primarily manifests in two aspects: on the one hand, the conflict of characteristics between different languages, and on the other hand, the competition between high-resource languages and medium- to low-resource languages. The conflict of language characteristics occurs when the model erroneously generalizes features from one language to others, leading to the incorrect application of syntactic, lexical, or other linguistic properties. For instance, English has explicit tense variations, while Chinese lacks tense markers; in Spanish, nouns and adjectives must agree in gender and number, a rule that does not exist in English. Furthermore, different languages may define the same entity category differently, which can result in classification errors for certain languages. For example, in English, "President" is typically categorized as "Title," whereas in Chinese, the corresponding term may be categorized as "Person." 
The competition between high-resource languages and medium- to low-resource languages refers to the fact that, due to differences in data volume, the characteristics of medium- to low-resource languages are often overshadowed by those of high-resource languages. The features of high-resource languages may dominate the model's performance, thereby weakening the model's ability to adapt to other languages. This phenomenon of “language competition” leads to a decline in the model’s performance when recognizing entities in medium- and low-resource languages.

To address the above issues, we propose RetrieveAll, a universal multilingual dynamically enhanced NER framework. As illustrated in figure~\ref{main}, leveraging the plug-and-play feature of LoRA modules, the framework constructs an input-aware LoRA retrieval mechanism, enabling hybrid multilingual recognition and reasoning across multiple languages. With a modular design to decouple task-specific language features, RetrieveAll allows flexible addition and updating of LoRA modules as languages change, independently of other language modules, effectively mitigating language interference. As a result, the framework maintains high performance while significantly reducing time and resource consumption, demonstrating efficiency, generalizability, and dynamic adaptability.

To enable RetrieveAll to better adapt to medium- and low-resource languages while alleviating the issue of data scarcity, we propose cross-granularity knowledge augmented learning. During LoRA fine-tuning, we design hierarchical prompts to jointly retrieve semantic-level and entity-level information without introducing additional data resources. This approach guides deeper knowledge injection, facilitating a paradigm shift from prompt-guided inference to prompt-driven learning, significantly enhancing the model's adaptability to tasks in medium- and low-resource languages.

Our main contributions can be summarized as follows:

\begin{itemize}
\item We propose RetrieveAll, a dynamic and universal multilingual NER framework based on LoRA modules. By decoupling task-specific features across different languages, RetrieveAll effectively mitigates language interference and supports efficient scalability. 
\item We propose a cross-granularity knowledge augmented learning that, without relying on additional data resources, integrates multi-granularity information through prompt-based fusion. 
\item Experimental results demonstrate that the RetrieveAll outperforms various baselines across multiple datasets, particularly on the PAN-X dataset, where it achieves an average F1 improvement of 12.1\%, thereby fully validating its advantages.

\end{itemize}

\begin{figure*}[t]
  \centering
  \includegraphics[width=1\textwidth]{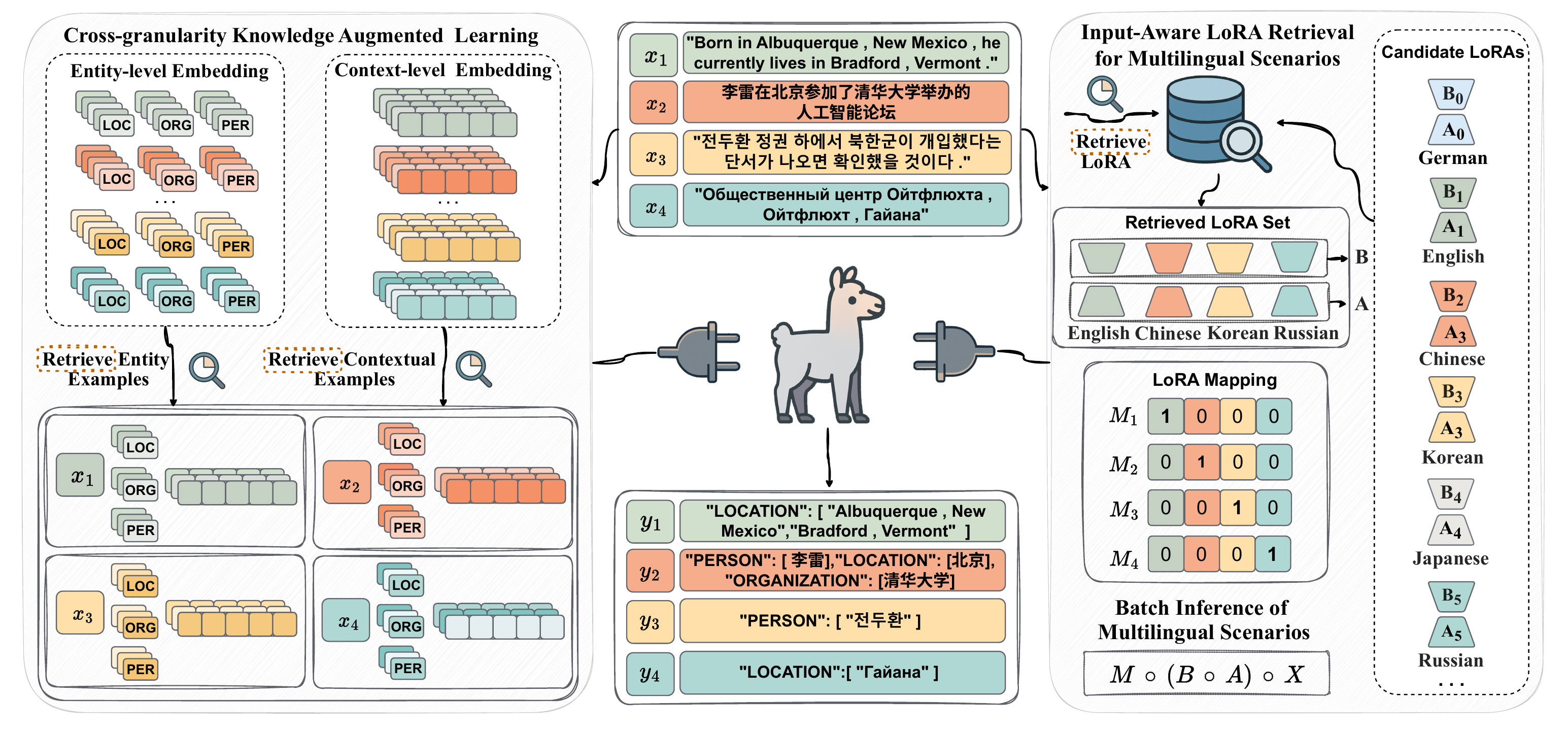}  
  \caption{RetrieveAll injects cross-granularity knowledge by retrieving entity-level and context-level examples via hierarchical prompts, while dynamically selecting and mapping the appropriate modules from a multilingual LoRA candidate pool based on the input language, enabling batch multilingual inference.}  
  \label{main}  
\end{figure*}

\section{Related Work}
Before the emergence of LLMs, most research treated NER as a sequence labeling task \citep{huang2020named, wu2022nflat}, typically assigning predefined labels (e.g., the BIO annotation scheme) to each token in the input text. 
LLMs have gained widespread acclaim for their outstanding performance across various NLP tasks. Driven by the advancement of LLMs, there has been a growing trend of applying seq2seq models to NER tasks \citep{zaratiana2023gliner,li2023codeie, keraghel2024recent}. Since LLMs were originally designed for text generation, their application to NER, a sequence labeling task, revealed certain limitations.

To address the above challenges, researchers have proposed various innovative approaches. Among them, prompt learning has emerged as an important research direction. The core idea of prompt learning is to better leverage LLMs by adding additional "prompts" \citep{liu2023pre, wang2022super, lu2023punifiedner}. Inspired by this, a wide range of prompting methods have been proposed in the literature. For example, GPT-NER \citep{wang2023gpt} reformulates sentences using special tokens, enabling LLMs to perform sequence labeling in a more natural text generation format. PromptNER \citep{ashok2023promptner} demonstrates how defining entity types in prompts allows LLMs to list entities with explanations. GL-NER \citep{zhu2024gl} employs a novel prompt template with label-injection instructions, enabling it to generate a "does not exist" signal when no entity is present. The above methods primarily focus on guiding model outputs by adjusting input prompts during the inference stage, whereas our approach emphasizes prompt-driven knowledge integration during the fine-tuning phase, demonstrating more pronounced advantages in knowledge guidance.

Additionally, prompting techniques have demonstrated great potential in unifying task paradigms. Prior work has designed generalized prompting methods to bring various single‐language tasks (e.g., multi‐class text classification \citep{lester2021power} and reading comprehension and QA \citep{khashabi2020unifiedqa}) under a single framework, thereby enabling transfer and sharing of model capabilities. However, these studies are largely confined to monolingual settings. In contrast, we focus on the issue of language interference in multilingual NER and propose the RetrieveAll framework, which aims to decouple task-specific features across different languages and effectively mitigate cross-language interference.

\section{Preliminaries}

Traditional full-parameter fine-tuning methods require high computational costs when fine-tuning LLMs, and their performance is limited in resource-constrained scenarios. To address this, LoRA \citep{Hu2021LoRALA} adds decomposable training matrices that are combined in parallel with the pre-trained parameters, fine-tuning only a small number of low-rank parameters. This reduces data dependency, lowers computational overhead, and enables efficient inference.

Specifically, given the pre-trained weights \( W_0 \in \mathbb{R}^{d \times d} \) of a sub-module of the LLM, LoRA introduces two trainable matrices \( A \) and \( B \) for low-rank adaptation, such that the weight update is \( \Delta W = BA \). Here, \( B \in \mathbb{R}^{d \times r} \), \( A \in \mathbb{R}^{r \times d} \), and \( r \) represents the rank of \( \Delta W \). The forward propagation formula is modified as follows:
\begin{equation}
y = W_0 x + \Delta W x = W_0 x + B A x, 
\end{equation}
where \( x \in \mathbb{R}^d \) is the input and \( y \in \mathbb{R}^d \) denotes the output.

\section{Methodology}

\subsection{Task reframing}
We decompose the multilingual NER task into several sub-tasks of NER for different languages.
Given an original multilingual LLM \( M \), its parameters are denoted as \( \theta_M \), and the set of \( k \) candidate LoRAs injected into model \( M \) is denoted as \( \theta_{P(M)} \), formally represented as:
\begin{equation}
\theta_{P(M)} = \left\{ \theta_{P(M)}^{1}, \theta_{P(M)}^{2}, \dots, \theta_{P(M)}^{k} \right\},
\end{equation}
where each \( \theta_{P(M)}^i \) is trained on the NER task for the specific language \( L_i \) and is composed of the matrices \( B_i \in \mathbb{R}^{d \times r} \) and \( A_i \in \mathbb{R}^{r \times d} \). The input of the multilingual NER task may come from any of the predefined \( k \) languages, or a combination of them. The input text sequence can be formulated as:
\begin{equation}
T_{\text{mix}} = \left\{ x \mid x \in \bigcup_{i=1}^{k} T_i \right\},
\end{equation}
where \( T_i \) represents the set of texts in language \( L_i \), and \( \cup \) denotes the union operation.

Given an input \( x \in T_{\text{mix}} \), without an explicit language label, the service process can be expressed as:
\begin{equation}
y = f_{\text{LoRA}}(x, \theta_M, g(\theta_{P(M)}, x)),
\end{equation}
where \( g(\theta_{P(M)}, x) \) denotes the input-aware LoRA retrieval process based on the input \( x \), which returns a set of retrieved LoRAs \( \theta_{P(M)}^i \) corresponding to the languages of the input \( x \). \( f_{\text{LoRA}}(x, \theta_M, \theta_{P(M)}^i) \) represents the LoRA composition process, which combines the retrieved LoRAs \( \theta_{P(M)}^i \) as a plug with the original parameters \( \theta_M \) of the LLM to generate the final output.

We believe that the design of RetrieveAll has the following advantages: (1) By independently training LoRA modules for specific languages, it enhances the model's focus on language-specific task features, effectively suppressing interference and conflicts between languages; (2) The lightweight parameter update mechanism based on LoRA allows the model to quickly converge even in environments with limited data resources, significantly reducing computational resource consumption and effectively mitigating the risk of overfitting in medium- and low- resource languages.; (3) When supporting new languages, it is sufficient to expand the LoRAs pool \( \theta_{P(M)} \) without adjusting the LLM parameters \( \theta_M \), enabling low-cost language expansion and greatly reducing the iteration costs of multilingual NER; (4) Through input-aware retrieval and dynamic combination of LoRAs, RetrieveAll can dynamically select suitable LoRAs based on the input language, thereby enhancing robustness in complex multilingual scenarios.

\subsection{Cross-granularity Knowledge Augmented Learning}
Contextual text and entity mentions are two key factors that significantly influence the performance of NER \citep{Jiang2024PICLPI}. Contextual text enhances the model's ability to understand entities through semantic associations, while entity mentions provide explicit markers that help the model accurately locate and classify entities. However, in NER tasks for some low- and medium-resource languages, models often struggle to fully leverage these two key factors due to insufficient training data.

Unlike previous studies that enhance retrieval-augmented generation (RAG) with external knowledge bases \citep{tan2023damo,lewis2020retrieval}, our goal is to improve the ability of model to process and learn from data itself, maximizing the extraction of valuable information without dependence on external resources. To achieve this, we propose a cross-granularity knowledge augmented learning approach. By incorporating contextual and entity examples into the training of the LoRA module, our method fully exploits valuable information inherent in existing training data. Leveraging knowledge at multiple granularities enables the LoRA module to accurately capture task-specific features of NER, significantly enhancing its adaptability and generalization to low- and medium-resource language scenarios.

Formally, given an original text sequence \( x \), we convert \( x \) into an input sequence \( x_{\text{input}} \) by prefixing it with a series of prompts at different granularities, as follows:
{\fontsize{9.5pt}{\baselineskip}\selectfont
\begin{equation}
x_{\text{input}} = (P_e, s_{e_1}, \dots, s_{e_n}, P_c, s_{c_1}, \dots, s_{c_m}, P_t, x),
\end{equation}
}
where, \( P_e \)  is the prompt text for entity examples, indicating that the following content lists example entities corresponding to the target entity types; \( s_{e_i} \) refers to specific entity examples, serving to clarify representative entities for each entity type; \( P_c \) is the prompt text for contextual examples, signaling that the subsequent content consists of context examples; \( s_{c_i} \) denotes specific contextual examples, each comprising a sample sentence along with its corresponding entity types and annotated entities; \( P_t \) is the prompt text that guides the output format, indicating that the following sentence is the original input from which entities are to be extracted. Then the target sequence \( y \) is as follows:
{\fontsize{9.5pt}{\baselineskip}\selectfont
\begin{equation}
y = ((ent_{p_1} : [ent_{e_1}]), \dots, (ent_{p_n} : [ent_{e_n}])),
\end{equation}
}
where \( ent_{p_i} \) is the entity type extracted from the input sequence; and \( ent_{e_i} \) is the ground truth text corresponding to the entity type \( ent_{p_i} \). Detailed input–output examples are shown in Appendix~\ref{sec2:appendix}.

In cross-granularity knowledge injection, the quality of contextual and entity examples is crucial to enhancing the effectiveness of knowledge injection. Representative entity examples provide models with clear entity boundaries and salient category features, significantly improving entity recognition and classification accuracy. For instance, distinguishing finer-grained types such as cities versus natural landmarks within the “Location” category. High-quality contextual examples establish robust mappings between entities and their categories across diverse contexts, effectively disambiguating cases such as “Apple” in technological versus everyday usage.

During the model’s knowledge injection phase, we fully leverage all training samples in the dataset to construct the examples required for multi-granularity knowledge augmentation. Given the training dataset for \( N \) samples,
\(
D = \{(x_{i}, y_{i})\}_{i=1}^N,
\)
where each sample consists of a text sequence \(x_{i}\) and its corresponding entity annotation sequence
\(
y_{i} = \{(e_{i}^{j},\,y_{i}^{j})\}_{j=1}^{l},
\)
\(y \in \mathcal{Y} \) is a list of \(l\) tuples, where \(e\) and \(y\) denote the contained entity and its entity type, and $\mathcal{Y}$ is a predefined all entity type set.

We employ LLMs as encoders to ensure consistent knowledge representation through their powerful semantic encoding capabilities. Specifically, for each sample \(D_i\), the LLMs encoder generates  entity-level representations \( \{E_{i}^{j}\}_{j=1}^{l} \in\mathbb{R}^d \) and context-level representations \( C_i \in \mathbb{R}^d \) of the same dimension. We obtain the semantic representation of the entity through the following formula: 
\begin{equation}
E_i^j = \text{Encoder}(e_i^j) \in \mathbb{R}^d,
\end{equation}
where \( d \) is the representation dimension, which generally depends on the encoder used. We obtain its contextual semantic representation using the following formula: 
\begin{equation}
C_i = \text{Encoder}(x_i) \in \mathbb{R}^d
\end{equation}
After obtaining all the entity representations \( E \) and context representations \( C \) of the training samples, we then select the most appropriate contextual and entity examples for each sample \(D_i\) based on cosine similarity:
\begin{equation}
\text{Sim}(D_i, E_j) = \frac{D_i \cdot E_j}{\|D_i\| \|E_j\|}
\end{equation}
\begin{equation}
\text{Sim}(D_i, C_j) = \frac{D_i \cdot C_j}{\|D_i\| \|C_j\|}
\end{equation}

Based on the dynamic similarity thresholds \( \tau_e \) and \( \tau_c \), we select the top-k entity examples with similarity higher than \( \tau_e \) for each entity type in the dataset of the training sample \(D_i\) (excluding the labeled entities in \(D_i\)), and select the top-k contextual examples with similarity higher than \( \tau_c \) (excluding \(D_i\)).

\begin{table*}[!t]
\begin{center}
\resizebox{\textwidth}{!}{
\begin{tabular}{lcccccccccccccc}
\toprule
\multirow{2}{*}{\textbf{Model}} & \multicolumn{8}{c}{\textbf{PAN-X}} & \multicolumn{6}{c}{\textbf{MultiCoNER}} \\
\cmidrule(lr){2-9} \cmidrule(lr){10-15}
& \textbf{en} & \textbf{es} & \textbf{fr} & \textbf{ru} & \textbf{de} & \textbf{zh} & \textbf{ja} & \textbf{ko}
& \textbf{en} & \textbf{es} & \textbf{ru} & \textbf{de} & \textbf{zh} & \textbf{ko} \\
\midrule
XLM-RoBERTa & 88.1 & 86.5 & 85.4 & 86.3 & 83.1 & 78.3 & 75.6 & 82.0 & 58.9 & 54.8 & 55.9 & 60.6 & 62.6 & 52.0 \\
LS-unlLaMA & 91.7 & 87.0 & 83.3 & 78.6 & 84.6 & 83.3 & 71.7 & 71.9 & 83.1 & 79.8 & 77.8 & 81.4 & 73.8 & 71.6 \\
PromptNER & 81.7 & 79.6 & 73.5 & 73.8 & 71.9 & 72.1 & 70.8 & 73.5 & 79.5 & 75.6 & 76.5 & 67.6 & 70.8 & 72.4 \\
GPT-NER & 75.2 & 72.8 & 71.6 & 63.5 & 72.0 & 72.4 & 71.5 & 72.1 & 71.7 & 67.9 & 58.2 & 63.1 & 61.2 & 62.5 \\
CascadeNER & 91.0 & 85.2 & 87.2 & 86.8 & 82.8 & 87.0 & 83.2 & 79.4 & \textbf{85.9} & \textbf{81.1} & \textbf{79.5} & 69.1 & \textbf{85.1} & \textbf{76.9} \\
\midrule
RetrieveAll (LLaMA3-8B, base) & 90.1 & 94.7 & 94.6 & 91.5 & 92.9 & 83.9 & 78.8 & 89.2 & 64.5 & 61.2 & 51.6 & 64.7 & 60.4 & 58.0 \\
RetrieveAll (LLaMA3-8B, CKAL) & 96.6 & \textbf{98.7} & \textbf{98.3} & 98.0 & \textbf{98.1} & 93.3 & 89.2 & 98.0 & 81.6 & 78.4 & 74.3 & 84.3 & 79.6 & 65.8 \\
RetrieveAll (Qwen2.5-7B, CKAL) & 96.4 & 98.6 & 98.1 & 97.8 & 98.0 & 93.3 & 89.7 & 97.7 & 79.1 & 77.1 & 73.7 & 82.5 & 81.2 & 64.8 \\
RetrieveAll (Qwen2.5-14B, CKAL) & \textbf{96.7} & 98.6 & 98.2 & \textbf{98.1} & \textbf{98.1} & \textbf{93.7} & \textbf{90.4} & \textbf{98.1} & 81.7 & 79.2 & 76.7 & \textbf{84.8} & 83.6 & 69.6 \\
\bottomrule
\end{tabular}
}
\end{center}
\caption{F1 scores (\%) of different models on PAN-X and MultiCoNER datasets. RetrieveAll (Base) denotes the results obtained without cross-granularity knowledge augmented learning, directly using the same inference approach. RetrieveAll (CKAL) denotes the results obtained with cross-granularity knowledge augmented learning. The best results are highlighted in bold.}
\label{main}
\end{table*}

\subsection{Input-Aware LoRA Retrieval for Multilingual Scenarios}
After identifying the top \( k \) most relevant contextual examples, we infer the language of the input sequence \( x_i\) by calculating the most frequent language label from the set \( \{l_1, l_2, \dots, l_k\} \) corresponding to these examples. This process can be expressed as:
{\fontsize{9.2pt}{\baselineskip}\selectfont
\begin{equation}
g(\theta_{P(M)}, x_i) = \{ \theta_{P(M)}^i , i=mode(l_1, l_2, \dots, l_k) \},
\end{equation}
}
where \( mode \) represents the mode function.

\subsection{Batch Inference of Multilingual Scenarios}
After retrieving the corresponding language LoRA modules \( \theta_{P(M)} \) for the input \( x_i \), we integrate these LoRA modules with the LLM parameters \( \theta_M \). Since mixed language scenarios may be involved, batch inference  \citep{zhao2024loraretriever} requires the use of multiple LoRA modules.

Given the batch input sequence
\( X \in \mathbb{R}^{b \times l \times d} \), 
where \( b \) denotes the batch size, \( l \) the sequence length, and \( d \) the input sequence dimensionality. 
For each input sequence \( x_i \), we retrieve its corresponding LoRA \(\theta_{P(M)}^i \) and then merge all \(\theta_{P(M)}^i \) into a global set 
\( \theta_B \). Since retrieval results for different input sequences may overlap, we remove duplicates from \( \theta_B \) , yielding a set of 
\( \theta_B \) \( p \) unique LoRA modules, with \( p \leq b\ \). Let us denote the submodule parameters in the batched LoRA set \(\theta_B\) as \( A = \{A_1, A_2, \dots, A_p\}\),\( \quad B = \{B_1, B_2, \dots, B_p\}.\) Concatenating these along the module dimension within \(\theta_B\) yields \( A \in \mathbb{R}^{p \times r \times d} \), \( \quad B \in \mathbb{R}^{p \times d \times r}.\)

For each input sequence \(x_i\), we construct a \(p\) dimensional mapping vector \(M_i\), whose entries indicate the indices of the LoRAs corresponding to that input sequence in the global set \(\theta_B\). Stacking all such vectors yields the mapping matrix \( M \in \mathbb{R}^{b \times p} \). 

In multilingual scenarios, the batch inference process of LoRA mixture can be formulated as:
\begin{equation}
Y =  M \,\circ\, (B \,\circ\, A \,)\circ\, X,
\end{equation}
where \(Y \in \mathbb{R}^{b \times l \times d}\) denotes the batched output of a layer of multiple LoRAs, and the operator \(\circ\) is used for broadcasting \citep{wen2023batched}.

\begin{table*}[!t]
\begin{center}
\resizebox{\textwidth}{!}{
\begin{tabular}{lcccccccccccccc}
\toprule
\multirow{2}{*}{\textbf{Model}} & \multicolumn{8}{c}{\textbf{PAN-X}} & \multicolumn{6}{c}{\textbf{MultiCoNER}} \\
\cmidrule(lr){2-9} \cmidrule(lr){10-15}
& \textbf{en} & \textbf{es} & \textbf{fr} & \textbf{ru} & \textbf{de} & \textbf{zh} & \textbf{ja} & \textbf{ko}
& \textbf{en} & \textbf{es} & \textbf{ru} & \textbf{de} & \textbf{zh} & \textbf{ko} \\
\midrule
RetrieveAll (CKAL, zero-shot, joint) & 87.1 & 94.0 & 93.4 & 91.2 & 90.9 & \textbf{78.7} & \textbf{73.7} & \textbf{86.5} & 79.4 & 75.3 & 69.4 & 80.9 & \textbf{63.5} & \textbf{64.9} \\
RetrieveAll (CKAL, zero-shot, monolingual) & \textbf{89.4} & \textbf{95.1} & \textbf{94.5} & \textbf{92.0} & \textbf{91.0} & 76.1 & 70.1 & 85.4 & \textbf{79.5} & \textbf{76.1} & \textbf{69.8} & \textbf{82.4} & 61.8 & 63.8 \\
\bottomrule
\end{tabular}
}
\end{center}
\caption{Zero-shot F1 Score (\%) comparison between cross-lingual joint training and monolingual training during cross-granularity knowledge augmented learning with RetrieveAll. The better results are highlighted in bold.}
\label{mono}
\end{table*}

\begin{table*}[!t]
\begin{center}
\resizebox{\textwidth}{!}{
\begin{tabular}{lcccccccccccccc}
\toprule
\multirow{2}{*}{\textbf{Model}} & \multicolumn{8}{c}{\textbf{PAN-X}} & \multicolumn{6}{c}{\textbf{MultiCoNER}} \\
\cmidrule(lr){2-9} \cmidrule(lr){10-15}
& \textbf{en} & \textbf{es} & \textbf{fr} & \textbf{ru} & \textbf{de} & \textbf{zh} & \textbf{ja} & \textbf{ko}
& \textbf{en} & \textbf{es} & \textbf{ru} & \textbf{de} & \textbf{zh} & \textbf{ko} \\
\midrule
RetrieveAll (CKAL, only entity-level examples) & 94.1 & 97.4 & 96.8 & 95.8 & 95.5 & 89.1 & 83.7 & 95.3 & 80.7 & 77.5 & 73.2 & 83.5 & 78.0 & 56.1 \\
RetrieveAll (CKAL, only context-level examples) & \textbf{96.9} & 98.6 & \textbf{98.4} & 97.9 & 98.0 & 93.2 & 89.2 & 97.9 & 81.3 & \textbf{78.6} & 74.2 & 83.8 & 79.2 & 64.7 \\
RetrieveAll (CKAL) & 96.6 & \textbf{98.7} & 98.3 & \textbf{98.0} & \textbf{98.1} & \textbf{93.3} & \textbf{89.5} & \textbf{98.0} & \textbf{81.6} & 78.4 & \textbf{74.3} & \textbf{84.3} & \textbf{79.6} & \textbf{65.8} \\
\bottomrule
\end{tabular}
}
\end{center}
\caption{Ablation results (\%) of entity-level and context-level examples in RetrieveAll during cross-granularity knowledge augmented learning. The better results are highlighted in bold.}
\label{CKAL}
\end{table*}

\begin{table*}[!t]
\begin{center}
\resizebox{0.8\textwidth}{!}{
\begin{tabular}{lcccccccccccccc}
\toprule
\multirow{2}{*}{\textbf{Model}} & \multicolumn{8}{c}{\textbf{PAN-X}} & \multicolumn{6}{c}{\textbf{MultiCoNER}} \\
\cmidrule(lr){2-9} \cmidrule(lr){10-15}
& \textbf{en} & \textbf{es} & \textbf{fr} & \textbf{ru} & \textbf{de} & \textbf{zh} & \textbf{ja} & \textbf{ko}
& \textbf{en} & \textbf{es} & \textbf{ru} & \textbf{de} & \textbf{zh} & \textbf{ko} \\
\midrule
RetrieveAll & 92.5 & 95.9 & 94.2 & 99.1 & 96.5 & 97.9 & 98.0 & 99.7 & 71.9 & 90.8 & 98.4 & 89.7 & 96.9 & 98.8 \\
\bottomrule
\end{tabular}
}
\end{center}
\caption{Accuracy (\%) of input-aware LoRA retrieval for different languages.}
\label{lora}
\end{table*}

\section{Experiments}
\subsection{Experiment Setting}
\paragraph{Implementation}
 In the RetrieveAll framework, we use LLaMA3-8B \footnote{\url{https://huggingface.co/meta-llama/Meta-Llama-3-8B}} \citep{grattafiori2024llama}, Qwen2.5-7B and Qwen2.5-14B  \footnote{\url{https://huggingface.co/Qwen}} \citep{yang2024qwen2}as base models, by default choosing LLaMA3-8B, and leverage LLM2Vec \citep{behnamghader2024llm2vec} as the encoder to encode each entity and its corresponding context in the dataset. To prevent information leakage, we do not use any data from the test set throughout the entire process. We set the similarity thresholds \( \tau_e \) and \( \tau_c \) to 0.65 and 0.7, and the parameter \( k \) to 5. By default, we select the top-5 contextual examples and entity examples with similarity scores greater than the thresholds. During inference, we adopt the same input format as used in the cross-granularity knowledge augmented learning stage. For evaluation, we primarily use the F1 score, as it is widely regarded as the most robust and effective metric for NER tasks \citep{li2019dice}. To ensure the reliability of the results, each model was independently evaluated over 10 runs, and the average performance was reported at a 95\% confidence level. See the Appendix~\ref{sec1:appendix} for detailed experimental settings.

\paragraph{Dataset}
We conducted experiments on two widely used multilingual datasets, MultiCoNER \citep{malmasi2022multiconer} and PAN-X \citep{pan2017cross}. MultiCoNER covers 6 entity types across 11 languages, while PAN-X includes 3 entity types spanning up to 282 languages. We selected 6 languages from MultiCoNER and 8 languages from PAN-X to evaluate the multilingual NER capabilities of RetrieveAll.

\paragraph{Baselines}
We selected five representative and competitive methods, including both supervised and LLM-based approaches, as baselines. For supervised methods, we adopted XLM-RoBERTa \citep{conneau2019unsupervised} in the MultiCoNER dataset. For LLM-based methods, we included GPT-NER \citep{wang2023gpt} based on GPT-3.5; LS-unLLaMA \citep{li2023label} based on LLaMA2-7B; PromptNER based on GPT-4 \citep{ashok2023promptner} ;and CascadeNER \citep{luo2024geic} which achieved SOTA results on several languages in both the MultiCoNER and PAN-X datasets.

\subsection{Main Results}

As shown in Table~\ref{main}, even without incorporating cross-granularity knowledge augmented learning (CKAL), RetrieveAll (LLaMA3-8B, Base) still demonstrates significant advantages on the PAN-X dataset, notably surpassing previous baselines on languages such as Spanish, French, Russian, and German. This result validates the effectiveness of the dynamic LoRA modules in decoupling language-specific features, significantly alleviating feature conflicts in multilingual scenarios, and ensuring robust adaptation to multilingual tasks.


After incorporating CKAL, RetrieveAll(LLaMA3-8B, CKAL) achieves significant performance gains across all languages on the PAN-X dataset. Compared to the strongest baseline for each language, the improvements are as follows: English by 5.3\%, Spanish by 13.4\%, French by 12.7\%, Russian by 12.9\%, German by 16.0\%, Chinese by 7.2\%, Japanese by 7.2\%, and Korean by 19.5\%. This method employs a prompt-guided knowledge injection mechanism, effectively integrating entity-level and context-level examples, achieving balanced multilingual performance optimization without relying on additional data resources.

To validate the adaptability of RetrieveAll, we conducted comparative experiments using base models of different scales (LLaMA3-8B and Qwen2.5-7B/14B). The results show that at similar parameter scales, LLaMA3-8B and Qwen2.5-7B deliver nearly identical performance on PAN-X, demonstrating that RetrieveAll can generalize across various base models without relying on a specific architecture. Furthermore, when scaling Qwen2.5 from 7B to 14B, the overall F1 score fluctuates by less than 1\%, further confirming RetrieveAll’s insensitivity to base model size and its flexibility in deployment.

Although RetrieveAll’s overall performance on the MultiCoNER dataset is slightly lower than on PAN-X, it remains highly competitive. RetrieveAll (LLaMA3-8B, CKAL) achieves F1 improvements of 26.5\% on English and 30.3\% on German over the base version, fully validating the effectiveness of cross-granularity knowledge augmented learning in capturing complex semantics; RetrieveAll (Qwen-14B, CKAL) attains a peak F1 score of 84.8\% on German. The limitations of RetrieveAll on the MultiCoNER dataset stem from two main factors: first, CKAL relies on context-level semantic alignment, but MultiCoNER contains a large proportion of short texts, making it difficult to deeply capture entity–context relationships; second, the dataset’s entity types follow a long-tail distribution and exhibit high sensitivity to noise, causing the dynamic LoRA modules to be easily disrupted by noise when adapting to language-specific features, thereby constraining the model’s generalization ability.

\subsection{Ablation Study and Discussion}
\paragraph{Decoupling Capability of Multilingual NER}
As shown in Table~\ref{mono}, to validate the multilingual feature decoupling capability of RetrieveAll, we conducted ablation experiments by removing the entity-level and context-level examples in CKAL, retaining only the dynamic LoRA modules for evaluation. The results show that monolingual training significantly outperforms joint training in most languages, confirming the effectiveness of RetrieveAll's decoupling mechanism. This demonstrates that dynamic LoRA effectively isolates task-specific language features, thereby mitigating competitive interference across languages. However, certain medium- and low-resource languages still face limitations due to inherent linguistic complexity, resulting in constrained performance.

\paragraph{Effect of Knowledge at Different Granularities}

As shown in Table~\ref{CKAL}, we conducted ablation experiments to compare the impact of different granularities of knowledge injection on model performance. The results confirm the necessity of cross-granularity knowledge integration across different languages. In the Korean NER task of MultiCoNER, the model that integrates both granularities of knowledge achieved a 17.3\% improvement compared to using only entity-level examples. RetrieveAll dynamically balances the different types of knowledge, maintaining performance ceilings in high-resource languages while achieving robust improvements in medium- and low-resource languages. The context-level examples serve as the core driver of performance improvement by enhancing the model’s semantic understanding through contextual modeling, while entity-level examples act as a complementary mechanism to assist in defining entity boundaries in medium- and low-resource languages.

\paragraph{Effect of Input-Aware LoRA Retrieval}
As shown in Table~\ref{lora}, experimental results show that the input-aware LoRA retrieval mechanism achieves accuracy rates of 99.1\% and 99.7\% on the Russian and Korean tasks of the PAN-X dataset, and 96.9\% and 98.8\% on the Chinese and Korean tasks of the MultiCoNER dataset, respectively. The input-aware dynamic LoRA retrieval mechanism ensures stable performance in high-resource language scenarios while achieving efficient generalization in medium- and low-resource languages, validating the strong robustness of the framework in multilingual settings.

\paragraph{Example Augmentation at Different Stages}
\begin{figure}[t]
  \centering
  \includegraphics[width=\columnwidth]{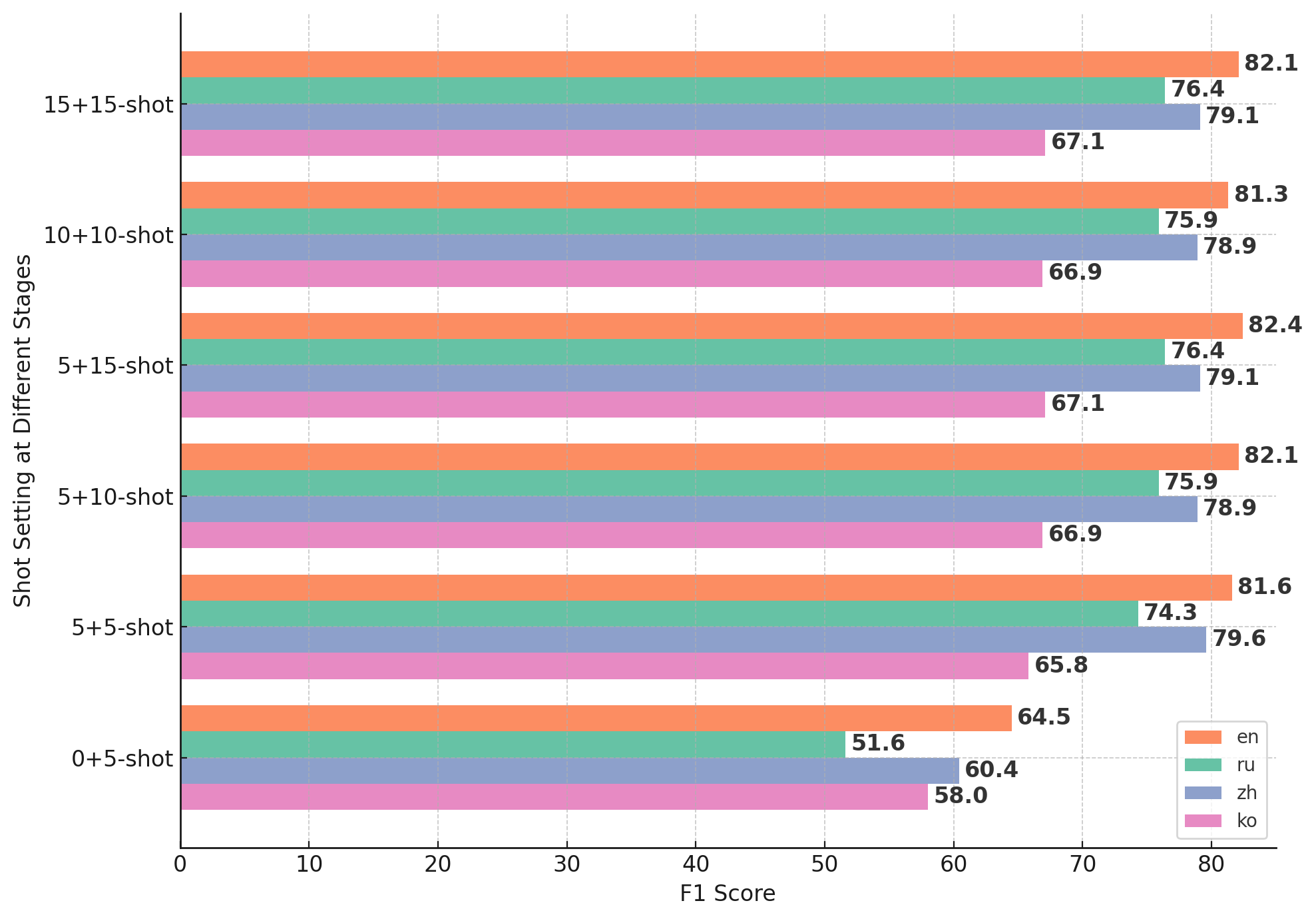}  
  \caption{Comparative analysis of the impact of example augmentation at different stages on RetrieveAll's performance.}  
  \label{shot}  
\end{figure}
We compared the impact of example counts between the training and inference stages on the MultiCoNER dataset. As shown in Figure~\ref{shot}, After incorporating CKAL during training and increasing the example count from 0+5-shot to 5+5-shot, the F1 scores for English, Russian, Chinese, and Korean improved by 26.5\%, 44.0\%, 31.8\%, and 13.4\%; in contrast, increasing inference examples from 5+5-shot to 5+15-shot produces only marginal changes of 1.0\%, 2.2\%, –0.7\%, and 2.0\%. These findings demonstrate that knowledge injection during the CKAL stage is far more critical than example expansion at inference, and in low-resource scenarios, priority should be given to strengthening training-stage knowledge injection.

\section{Conclusion}
This paper investigates key challenges in multilingual NER, including severe language interference, limited performance on medium- and low-resource languages, and the poor scalability of traditional approaches. To address these issues, we propose RetrieveAll, a universal framework based on dynamic LoRA modules. In addition, we introduce a cross-granularity knowledge augmentation, which also facilitates a paradigm shift from "prompt-guided inference" to "prompt-driven learning."

\section*{Limitations}
Although RetrieveAll’s dynamic LoRA modules significantly enhance system flexibility in multilingual high-concurrency scenarios through input-aware retrieval and batch inference mechanisms, they can also introduce additional computational latency; moreover, when target language resources are extremely scarce and retrievable examples are insufficient, relying solely on internal examples may fail to capture language characteristics adequately, thereby limiting the effectiveness of cross-granularity knowledge augmentation.

\bibliography{custom}

\newpage
\appendix
\section{Reproducibility and Robustness Evaluation of RetrieveAll
}
\label{sec1:appendix}

\begin{table*}[htbp]
  \centering
  \resizebox{\textwidth}{!}{%
\begin{tabular}{l c c c c c}
\toprule
\textbf{Model} & \textbf{Languages} & \textbf{AVG. F1} & \textbf{Std.\ Dev.} & \textbf{95\% CI Lower} & \textbf{95\% CI Upper} \\
\midrule
\multirow{7}{*}{\centering RetrieveAll (LLaMA3-8B, base)}  
  & en & 90.1 & 0.60 & 89.73 & 90.47 \\
  & es & 94.7 & 0.40 & 94.45 & 94.95 \\
  & fr & 94.6 & 0.50 & 94.29 & 94.91 \\
  & ru & 91.5 & 0.70 & 91.07 & 91.93 \\
  & de & 92.9 & 0.60 & 92.53 & 93.27 \\
  & zh & 83.9 & 1.10 & 83.22 & 84.58 \\
  & ja & 78.8 & 1.30 & 77.99 & 79.61 \\
  & ko & 89.2 & 0.80 & 88.71 & 89.70 \\
\midrule
\multirow{8}{*}{RetrieveAll (LLaMA3-8B, CKAL)}
  & en & 96.6 & 0.40 & 96.35 & 96.85 \\
  & es & 98.7 & 0.30 & 98.51 & 98.89 \\
  & fr & 98.3 & 0.50 & 97.99 & 98.61 \\
  & ru & 98.0 & 0.60 & 97.63 & 98.37 \\
  & de & 98.1 & 0.30 & 97.91 & 98.29 \\
  & zh & 93.3 & 0.90 & 92.74 & 93.86 \\
  & ja & 89.2 & 1.20 & 88.46 & 89.94 \\
  & ko & 98.0 & 0.50 & 97.69 & 98.31 \\
\midrule
\multirow{8}{*}{RetrieveAll (Qwen2.5-7B, CKAL)}
  & en & 96.4 & 0.50 & 96.09 & 96.71 \\
  & es & 98.6 & 0.60 & 98.23 & 98.97 \\
  & fr & 98.1 & 0.40 & 97.85 & 98.35 \\
  & ru & 97.8 & 0.70 & 97.37 & 98.23 \\
  & de & 98.0 & 0.50 & 97.69 & 98.31 \\
  & zh & 93.3 & 1.00 & 92.68 & 93.92 \\
  & ja & 89.7 & 0.90 & 89.14 & 90.26 \\
  & ko & 97.7 & 0.60 & 97.33 & 98.07 \\
\midrule
\multirow{8}{*}{RetrieveAll (Qwen2.5-14B, CKAL)}
  & en & 96.7 & 0.50 & 96.39 & 97.01 \\
  & es & 98.6 & 0.40 & 98.35 & 98.85 \\
  & fr & 98.2 & 0.60 & 97.83 & 98.57 \\
  & ru & 98.1 & 0.50 & 97.79 & 98.41 \\
  & de & 98.1 & 0.40 & 97.85 & 98.35 \\
  & zh & 93.7 & 0.80 & 93.21 & 94.20 \\
  & ja & 90.4 & 1.00 & 89.78 & 91.02 \\
  & ko & 98.1 & 0.70 & 97.67 & 98.53 \\
\bottomrule
\end{tabular}
}
\caption{Average F1 score, standard deviation, and 95\% confidence interval of four RetrieveAll variants (LLaMA3-8B (base), LLaMA3-8B (CKAL), Qwen2.5-7B (CKAL), and Qwen2.5-14B (CKAL)) across eight languages of the PAN-X dataset.
}
\label{tab:multilingual_top4}
\end{table*}

\begin{table*}[htbp]
  \centering
  \resizebox{\textwidth}{!}{%
\begin{tabular}{lcccccc}
\toprule
Model & Language & AVG. F1 & Std. Dev. & 95\% CI Lower & 95\% CI Upper \\
\midrule
\multirow{6}{*}{RetrieveAll (LLaMA3-8B, base)}
  & en & 64.5 & 1.50 & 63.57 & 65.43 \\
  & es & 61.2 & 1.20 & 60.46 & 61.94 \\
  & ru & 51.6 & 1.80 & 50.49 & 52.71 \\
  & de & 64.7 & 1.40 & 63.83 & 65.57 \\
  & zh & 60.4 & 1.60 & 59.41 & 61.39 \\
  & ko & 58.0 & 2.00 & 56.76 & 59.24 \\
\midrule
\multirow{6}{*}{RetrieveAll (LLaMA3-8B, CKAL)}
  & en & 81.6 & 1.00 & 80.98 & 82.22 \\
  & es & 78.4 & 0.80 & 77.90 & 78.90 \\
  & ru & 74.3 & 1.20 & 73.56 & 75.04 \\
  & de & 84.3 & 0.90 & 83.74 & 84.86 \\
  & zh & 79.6 & 0.70 & 79.17 & 80.03 \\
  & ko & 65.8 & 1.50 & 64.87 & 66.73 \\
\midrule
\multirow{6}{*}{RetrieveAll (Qwen2.5-7B, CKAL)}
  & en & 79.1 & 1.30 & 78.30 & 79.90 \\
  & es & 77.1 & 1.00 & 76.48 & 77.72 \\
  & ru & 73.7 & 0.90 & 73.14 & 74.26 \\
  & de & 82.5 & 1.10 & 81.82 & 83.18 \\
  & zh & 81.2 & 0.80 & 80.70 & 81.70 \\
  & ko & 64.8 & 1.40 & 63.93 & 65.67 \\
\midrule
\multirow{6}{*}{RetrieveAll (Qwen2.5-14B, CKAL)}
  & en & 81.7 & 1.10 & 81.02 & 82.38 \\
  & es & 79.2 & 0.90 & 78.64 & 79.76 \\
  & ru & 76.7 & 1.00 & 76.08 & 77.32 \\
  & de & 84.8 & 1.30 & 84.00 & 85.60 \\
  & zh & 83.6 & 0.80 & 83.10 & 84.10 \\
  & ko & 69.6 & 1.20 & 68.86 & 70.34 \\
\bottomrule
\end{tabular}
}
\caption{Average F1 score, standard deviation, and 95\% confidence interval of four RetrieveAll variants (LLaMA3-8B (base), LLaMA3-8B (CKAL), Qwen2.5-7B (CKAL), and Qwen2.5-14B (CKAL)) across six languages of the MultiCoNER dataset.}
\label{tab:retrieveall-multiconer}
\end{table*}

In our experiments, each model underwent ten independent evaluations, with  the average F1 score reported in the main text; in Tables~\ref{tab:multilingual_top4} and~\ref{tab:retrieveall-multiconer}, we present the error bars and confidence intervals for RetrieveAll variant to offer a more intuitive depiction of the reproducibility and robustness of the performance results.

\section{RetrieveAll Input and Output Format}
\label{sec2:appendix}
\begin{figure}[t]
  \includegraphics[width=\columnwidth]{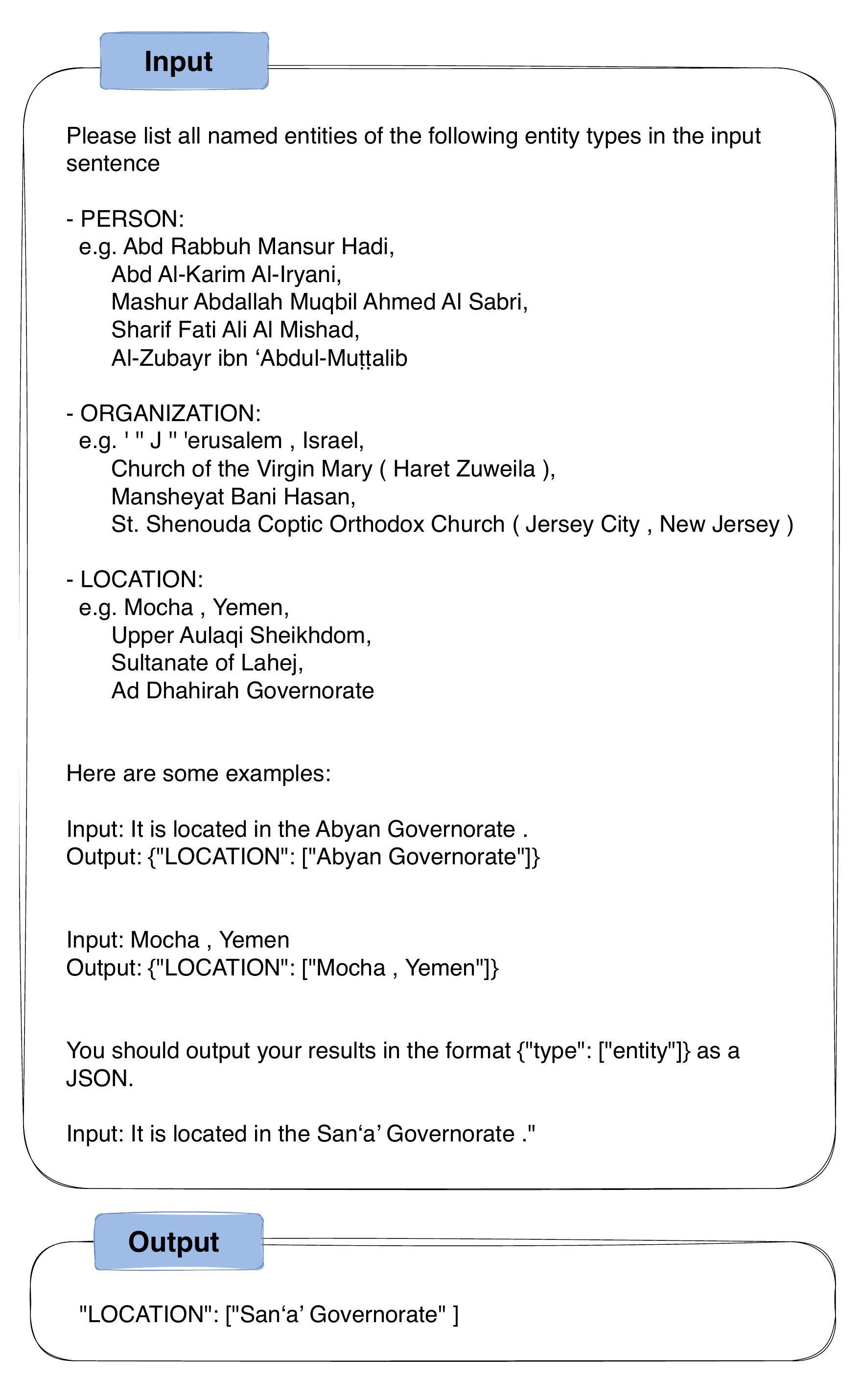}  
  \caption{The standardized input and output format of the RetrieveAll is illustrated using a specific data example.}  
  \label{example}  
\end{figure}
Figure~\ref{example} presents the specific standardized format of model input and output using a single data example, providing a more comprehensive illustration of its processing details.

\end{document}